\documentclass[letterpaper]{article} 
\usepackage{aaai23}  
\usepackage{times}  
\usepackage{helvet}  
\usepackage{courier}  
\usepackage[hyphens]{url}  
\usepackage{graphicx} 
\usepackage{natbib}  
\usepackage{caption} 
\usepackage{algorithm}
\usepackage{algorithmic}
\usepackage{dsfont}
\usepackage{amsmath}
\usepackage{amssymb}
\usepackage{booktabs}
\usepackage[figuresright]{rotating}
\usepackage{multirow}
\usepackage{url}
\usepackage{newfloat}
\usepackage{listings}
\DeclareCaptionStyle{ruled}{labelfont=normalfont,labelsep=colon,strut=off} 
\floatstyle{ruled}
\newfloat{listing}{tb}{lst}{}
\floatname{listing}{Listing}
\usepackage{xspace}
\newcommand{\ourapproach}{\textsc{T2G-Former}\xspace}
\title{\ourapproach: Organizing Tabular Features into Relation Graphs Promotes\\ Heterogeneous Feature Interaction}
\author{
    Jiahuan Yan\textsuperscript{\rm 1}\equalcontrib,
    Jintai Chen\textsuperscript{\rm 1}\equalcontrib,
    Yixuan Wu\textsuperscript{\rm 2},
    Danny Z. Chen\textsuperscript{\rm 3},
    Jian Wu\textsuperscript{\rm 4}\thanks{The corresponding author.}
}
\affiliations{
    \textsuperscript{\rm 1}College of Computer Science and Technology, Zhejiang University, Hangzhou, China\\
    \textsuperscript{\rm 2}School of Medicine, Zhejiang University, Hangzhou, China\\
    \textsuperscript{\rm 3}Department of Computer Science and Engineering, University of Notre Dame, Notre Dame, IN 46556, USA\\
    \textsuperscript{\rm 4}The First Affiliated Hospital, and Department of Public Health, Zhejiang University School of Medicine, Hangzhou, China\\

    jyansir@zju.edu.cn, jtchen721@gmail.com, wyx\_chloe@zju.edu.cn, dchen@nd.edu, wujian2000@zju.edu.cn
}

\usepackage{bibentry}

\begin{document}

\maketitle

\begin{abstract}
Recent development of deep neural networks (DNNs) for tabular learning has largely benefited from the capability of DNNs for automatic feature interaction. However, the heterogeneity nature of tabular features makes such features relatively independent, and developing effective methods to promote tabular feature interaction still remains an open problem. In this paper, we propose a novel \textit{Graph Estimator}, which automatically estimates the relations among tabular features and builds graphs by assigning edges between related features. Such relation graphs organize independent tabular features into a kind of graph data such that interaction of nodes (tabular features) can be conducted in an orderly fashion. Based on our proposed  \textit{Graph Estimator}, we present a bespoke Transformer network tailored for tabular learning, called \ourapproach, which processes tabular data by performing tabular feature interaction guided by the relation graphs. A specific \textit{Cross-level Readout} collects salient features predicted by the layers in \ourapproach across different levels, and attains global semantics for final prediction. Comprehensive experiments show that our \ourapproach achieves superior performance among DNNs and is competitive with non-deep Gradient Boosted Decision Tree models. The code and models are available at https://github.com/jyansir/t2g-former.
\end{abstract}

\section{Introduction} \label{intro}

Data in the form of table structures are ubiquitous in many fields, e.g., medical records~\cite{johnson2016mimic, hassan2020machine} and click-through rate (CTR) prediction~\cite{covington2016deep, song2019autoint}. It was observed that Gradient Boosted Decision Trees (GBDT)~\cite{chen2016xgboost, ke2017lightgbm, prokhorenkova2018catboost} were dominating models for tabular data tasks in machine learning and industrial applications. Due to big successes of deep neural networks (DNNs) in various fields, there has been increasing development of specialized DNNs for tabular data learning~\cite{popov2019neural, arik2021tabnet, wang2021dcn, gorishniy2021revisiting, chen2022danets}. Such studies either leveraged ensembling of neural networks~\cite{popov2019neural, arik2021tabnet, katzir2020net} to build \textit{differentiable tree models}, or explored diverse interaction approaches~\cite{guo2017deepfm, wang2017deep, song2019autoint, wang2021dcn, gorishniy2021revisiting, chen2022danets} to learn comprehensive features by fusing different tabular features.

\begin{figure*}[t]
\centering
\includegraphics[width=0.8\textwidth]{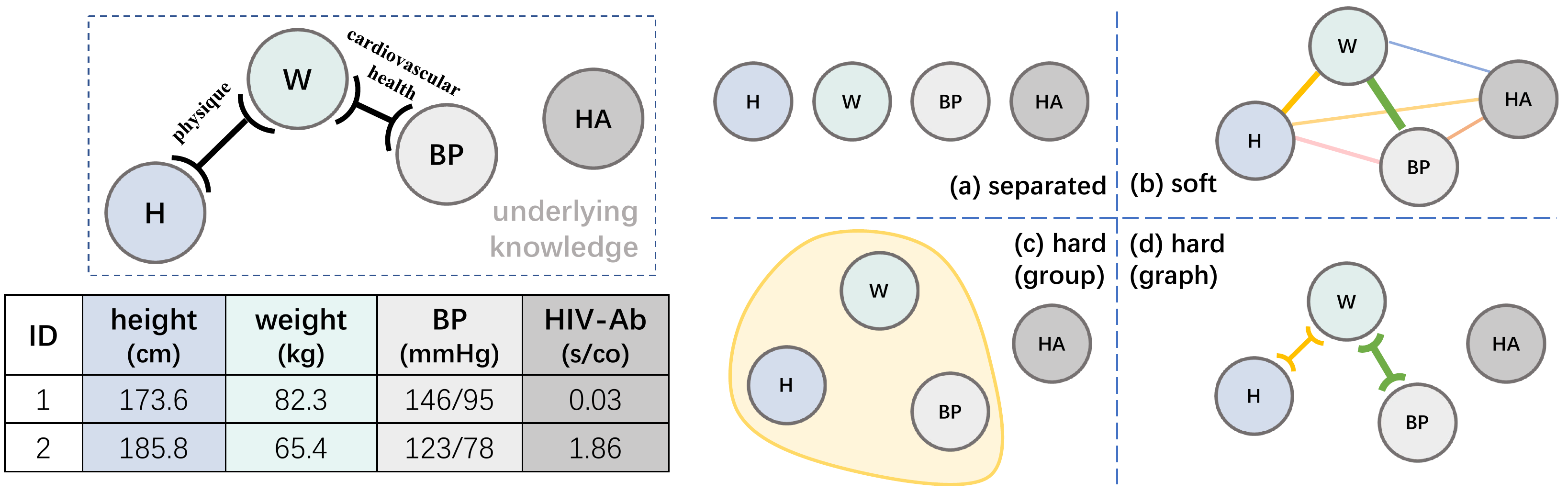} 
\caption{An example of medical data tables. The values in different columns are located in heterogeneous feature spaces. Underlying medical knowledge sparsely links feature pairs. (a) Original separated features without any interactions, which are often used in non-deep models. (b) Fully connected interactions by softly selecting all the features. (c) Selective interactions among grouped features by hard selection. (d) Selective interactions according to a weighted relation graph. ``BP'' denotes blood pressure; ``HIV-Ab'' indicates the level of HIV antibody.}
\label{fig-illustration}

\end{figure*}

However, different from images and texts, it is challenging for fusion-based models to handle tabular feature interaction due to the feature heterogeneity problem~\cite{borisov2021deep}. DANets~\cite{chen2022danets} suggested the ``selection \& abstraction'' principle that processes tabular data by first selecting and then interacting the selected features. Known neural feature selection schemes can be categorized into soft and hard versions. The soft selection essentially exerts fully connected interactions among features (see Fig.~\ref{fig-illustration}(b)), such as multiplicative interaction~\cite{guo2017deepfm}, feature crossing~\cite{wang2017deep, wang2021dcn}, and attention-based interaction~\cite{song2019autoint, huang2020tabtransformer, gorishniy2021revisiting}. However, tabular features by nature are heterogeneous, and fully connected interaction is a sub-optimal choice since it blindly fuses all features together. DANets~\cite{chen2022danets} performed hard selection by grouping correlative features and then constraining interactions among grouped features (Fig.\ref{fig-illustration}(c)). Although DANets achieved promising results, its feature selection operation cannot thoroughly address intra-group interactions (see Fig.~\ref{fig-illustration}(c)), and thus features assigned in a same group are indiscriminately fused, making the model inferiorly expressive.

There are numerous daily applications that exemplify the significance of selective interaction for heterogeneous tabular features. The left part of Fig.~\ref{fig-illustration} gives an example of a medical data table. 
Using underlying medical knowledge, a static graph can be formed to indicate relations of reasonable feature pairs. For instance, the relation of \textit{height} and \textit{weight} gives a probability representing a high-level semantic \textit{physique}. Also, the relation between \textit{weight} and \textit{blood pressure} (\textit{BP}) is likely to indicate a semantic \textit{cardiovascular health}. Besides, there might be some ``inert features'' that are unrelated to any other features, such as the features representing the \textit{level of HIV antibody} (\textit{HIV-Ab}).
In the right part, Fig.~\ref{fig-illustration}(a) presents the original tabular features whose relations are not specified, and higher-level semantics cannot be directly obtained if the feature relations are not determined. Fig.~\ref{fig-illustration}(b) illustrates the fully connected interactions of soft selection, which may introduce some noisy relations in feature fusion (e.g., the ``inert feature'' connects with the other features). Hard selection with a grouping operation (e.g., used in DANets) achieves partially selective interactions by grouping related features (see Fig.~\ref{fig-illustration}(c)), but is still likely to include noisy interactions. 
It can only group related features but fails to handle the feature relations within the same group. 
In Fig.~\ref{fig-illustration}(c), the grouping design can only put the features \textit{height}, \textit{weight}, and \textit{BP} together for mutual interactions, but cannot exclude the meaningless \textit{height}-\textit{BP} pair. It is intuitive that a precise health condition assessment can be made based on both the data-specific record values (e.g., 173.6 cm for \textit{height} in Fig.~\ref{fig-illustration}) and the underlying knowledge represented by the edges of the relation graph. For the first sample in the medical table (ID = 1), considering the values of \textit{height} and \textit{weight} jointly can suggest a symptom of overweight. Similarly, combining the values of \textit{weight} and \textit{BP} indicates a risk of cardiovascular problems. The second sample (ID = 2) directly indicates a risk of HIV infection solely based on the feature of \textit{HIV-Ab}. Hence, we argue that an ideal way to handle such complex decision processes is to build a graph with adaptive edge weights. The edge weights (represented by different colors and widths in Fig.~\ref{fig-illustration}(d)) indicate the strengths of relations based on specific feature values, and the static graph topology represents the underlying knowledge to constrain meaningful relations.

Inspired by the above observations, in this paper, we propose to build graphs for tabular features to guide feature interaction. We develop a novel \textit{Graph Estimator} (GE) for organizing independent tabular features into a feature relation graph (FR-Graph). Further, we present
a bespoke Transformer network for tabular learning, called \ourapproach, by stacking GE-incorporated blocks for selective feature interaction. GE models an FR-Graph by assembling (i) a static graph topology depicting underlying knowledge of the task and (ii) data-adaptive edge weights for graph edges. The static graph depicts the underlying knowledge (the relations of feature pairs), while the data-adaptive edge weights represent the strengths of relations based on specific feature values. Using the FR-Graph, we can effectively capture more subtle interactions which may be mishandled by grouping strategies (as shown in Fig.~\ref{fig-illustration}(c)). In our proposed \ourapproach, each layer employs the FR-Graph to transform layer input features into graph data, and heterogeneous feature interactions are performed in an orderly fashion based on the specification of graph edges. Besides, a special \textit{Cross-level Readout} collects salient features from each level and attains global tabular semantics for the final prediction.

The workflow of \ourapproach proceeds as follows. An FR-Graph, whose edges represent the static relations of features with data-adaptive weights (predicted by the GE module), guides the processing of the tabular feature interaction to predict higher-level features. Then another FR-Graph for higher-level tabular features is built to organize the feature interaction, and the process continues. \ourapproach can output comprehensive semantics from different feature levels by repeating the above process. The shared \textit{Cross-level Readout} is used to aggregate semantics from different feature levels, and takes all these features into consideration in the final prediction.

Overall, the main contributions of our work are as follows:

\begin{itemize}
    \item We first utilize feature relation graphs to handle heterogeneous feature interaction for tabular data, and propose a novel \textit{GE}  module for feature relation organization.
    \item We adapt feature relation graphs in the Transformer architecture, and build a specialized tabular learning Transformer \ourapproach for tabular classification and regression. 
    \item Comprehensive experiments show that \ourapproach consistently outperforms state-of-the-art tabular DNNs on many datasets, and is competitive with GBDTs.
\end{itemize}

\begin{figure*}[t]
\centering
\includegraphics[width=0.85\textwidth]{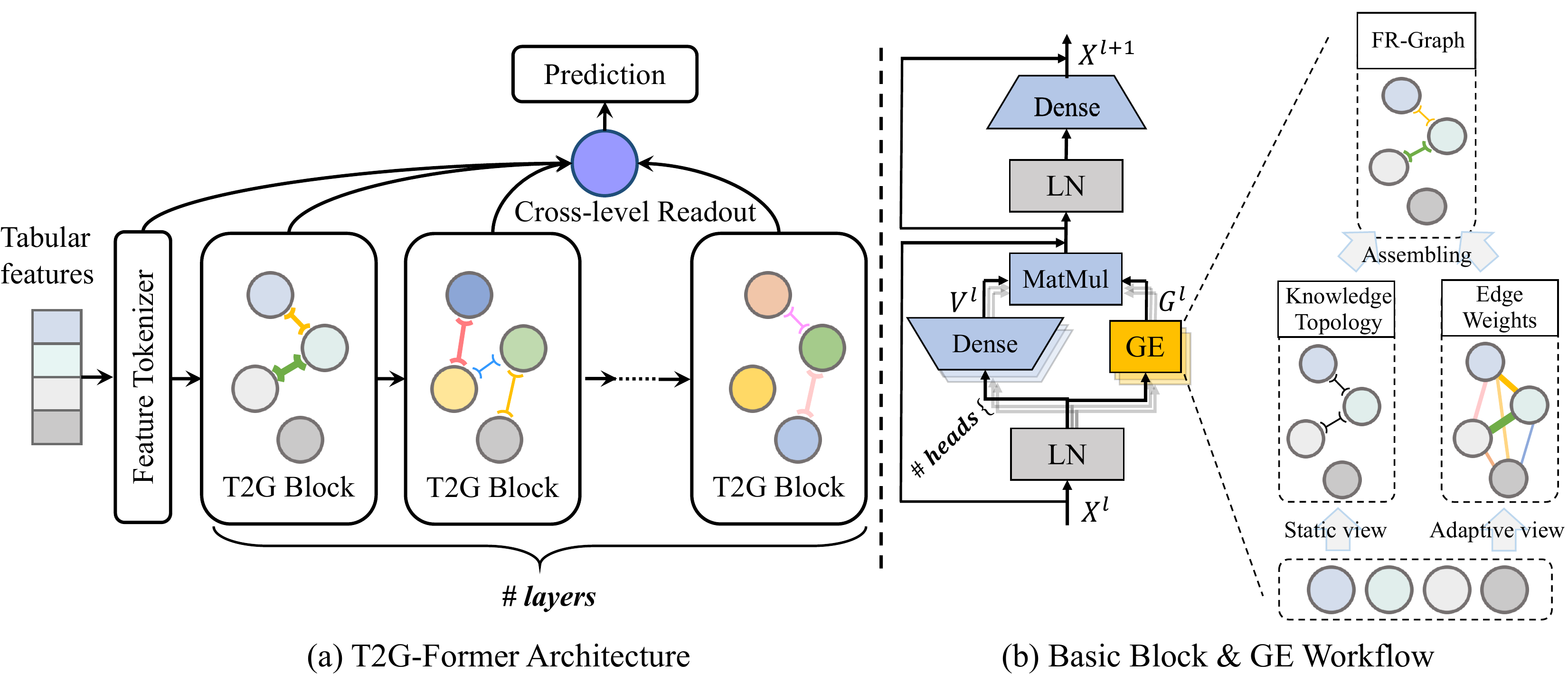} 

\caption{(a) The architecture of \ourapproach for tabular learning. Each T2G block builds an FR-Graph for a feature level and performs selective interaction. A global readout node collects salient features from each layer to form tabular semantics. (b) Illustrating a basic block in Sec.~\ref{bb} and GE in Sec.~\ref{ge}.}
\label{fig-t2g}

\end{figure*}

\section{Related Work}

\subsection{DNNs for Tabular Learning}

Tabular learning refers to machine learning applications on tabular data that conducts prediction based on categorical or continuous features~\cite{d:22}. Classical non-deep methods~\cite{li1984classification, friedman2001greedy, zhang2003learning, zhang2006learning, he2014practical} are prevalent choices for such tasks~\cite{anghel2018benchmarking}, especially the ensemble methods of decision trees, such as GBDT~\cite{friedman2001greedy}, XGBoost~\cite{chen2016xgboost}, LightBGM~\cite{ke2017lightgbm}, and CatBoost~\cite{prokhorenkova2018catboost}.

Compared to their shallow counterparts, DNNs enjoy strong abilities of automatic feature learning~\cite{thawani2021representing}, and hence offer a good potential to exploit hidden features. Recently, increasingly more studies applied DNNs to tabular data~\cite{guo2017deepfm, yang2018deep, song2019autoint, feng2018multi, hazimeh2020tree, popov2019neural, arik2021tabnet, chen2022danets}, which can be roughly categorized into \textbf{differentiable tree models} and \textbf{fusion-based models}.

\subsubsection{Differentiable Tree Models.} DNNs of this type~\cite{popov2019neural, arik2021tabnet, katzir2020net} were inspired by the successes of the ensemble tree frameworks~\cite{kontschieder2015deep, feng2018multi, yang2018deep}. NODE~\cite{popov2019neural} combined differentiable oblivious decision trees~\cite{lou2017bdt} with multi-layer hierarchical representations and achieved competitive performances as GBDT. TabNet~\cite{arik2021tabnet} employed an attention mechanism~\cite{vaswani2017attention} to sequentially select salient features for tree-like decision. Net-DNF~\cite{katzir2020net} introduced bias of a disjunctive normal form to select and aggregate feature subsets in each block. NODE and Net-DNF largely benefited from model ensembles but did not take advantage of the feature representation capability of DNNs~\cite{chen2022danets}. TabNet designed non-interactive transformer blocks for feature representation and selection without feature fusion. All these DNNs function as feature selectors and splitters, but neglect underlying interactions among tabular features.

\subsubsection{Fusion-based Models.} Fusion-based models~\cite{guo2017deepfm, song2019autoint, huang2020tabtransformer, wang2021dcn, gorishniy2021revisiting, chen2022danets} leveraged DNNs to fuse higher-level features via feature interaction. DeepFM~\cite{guo2017deepfm} performed multiplicative interaction on encoded features for CTR prediction. DCN~\cite{wang2017deep, wang2021dcn} combined DNNs with cross components to learn complex features with high-order interactions. Recently, attention module~\cite{vaswani2017attention} became a popular choice due to its interactive bias and remarkable performance~\cite{kenton2019bert, dosovitskiy2020image}. AutoInt~\cite{song2019autoint} used multi-head self-attention to interact low-dimension embedded features. TabTransformer~\cite{huang2020tabtransformer} directly transferred Transformer~\cite{vaswani2017attention} blocks to tabular data but neglected interaction between categorical features and continuous ones. FT-Transformer~\cite{gorishniy2021revisiting} addressed this problem by tokenizing these two types of features and processing them equally. DANets~\cite{chen2022danets} selected correlative tabular features and attentively fused the selected features into higher-level ones.

\subsection{Tabular Feature Interaction}

Most of the previous fusion-based work simply transferred successful neural architectures (e.g., MLP~\cite{guo2017deepfm}, self-attention~\cite{song2019autoint}, and Transformer~\cite{huang2020tabtransformer, gorishniy2021revisiting}) into tabular data and interacted features with soft selection. However, feature heterogeneity~\cite{borisov2021deep, popov2019neural} led to gap of inductive bias and made these models (which were designed for homogeneous data, e.g., images and texts) sub-optimal. DANets~\cite{chen2022danets} first adapted selective feature interaction by hard selection, constraining interactions in a feature group, and achieved promising results; but, relations of intra-group features were still not managed well. Hence, this paper proposes feature relation graphs and adapts them into a tailored Transformer network.

\section{Graph Estimator} \label{ge}

We propose \textit{Graph Estimator} (GE) (Fig.~\ref{fig-t2g}(b)) for automatically building \textbf{\textit{Feature Relation Graphs} (FR-Graphs)}, which treats tabular features as nodes in a graph and estimates the feature relations as edges. The GE design is inspired by knowledge graph completion (KGC)~\cite{shi2018open, wu2021graph} that might use semantical similarity of two entities to estimate their relation plausibility. A basic form to measure semantical similarity~\cite{nickel2011three} is:
\begin{equation}
    f_r(h,t) = h^T M_r t \label{bilinear},
\end{equation}
where $h, t \in \mathds{R}^n$ are an encoded head entity node and a tail one, and a learnable matrix $M_r \in \mathds{R}^{n \times n}$ represents relation $r$ in a knowledge graph (KG). Various following methods~\cite{yang2015embedding, trouillon2016complex, nickel2016holographic} followed this idea, which differed from one another solely in relation embeddings and score functions.

Different from KGC models that only compute static relation plausibility for entities, GE estimates the feature relations by a static underlying graph topology with data-adaptive edge weights. We take each tabular feature as a node, and first perform semantic matching to estimate the soft plausibility of pair-wise interactions between tabular features, which are referred to as \textbf{data-adaptive edge weights} in this section. Second, a \textbf{static knowledge topology} is learned based on tabular column semantics to preserve interactions of salient feature pairs. At the end, edge weights are assembled with the knowledge topology to form an FR-Graph.

\subsection{FR-Graph Structure Components}
To mine the relations among tabular features, we build FR-Graph by treating tabular features as graph node candidates and predicting the edges among them. The edges were yielded from two perspectives: adaptive edge weights representing data-specific information, and static edge topology for all the data representing the underlying knowledge. Note that some features are isolated from the FR-Graph if no other nodes connected with them.

\subsubsection{Adaptive Edge Weights.} Given two tabular feature embedding vectors $x_i, x_j \in \mathds{R}^n$ ($i,j \in \{1,2,\dots, N\})$, where $N$ is the number of input features (table columns), we evaluate their interaction plausibility using the following pair-wise score function:
\begin{equation}
    G_w[i,j] = g_w(f_i^h,f_j^t) = {f_i^h}^T \text{diag}(r) f_j^t \label{dg},
\end{equation}
\begin{equation}
    f_i^h = W^h x_i, f_i^t = W^t x_i, \
    \begin{cases}
    W^h \equiv W^t& \text{if symmetric},\\
    W^h \ne W^t& \text{if asymmetric},
    \end{cases}  \label{ft-dg}
\end{equation}
where two learnable parameters $W^h, W^t \in \mathds{R}^{m \times n}$ denote transformations for a head feature and a tail one, and $\text{diag}(r) \in \mathds{R}^{n \times n}$ is a diagonal matrix parameterized by learnable relation vectors $r \in \mathds{R}^n$ that semantically represent feature interaction relations. 
Here $W^h$ and $W^t$ share parameters if the pair-wise feature edge weights are symmetric (i.e., $G_w[i,j] \equiv G_w[j,i]$) and are parameter-independent in the asymmetric case (i.e., $G_w[i,j] \ne G_w[j,i]$). All bias vectors are omitted for notation brevity. Consequently, the adaptive weight scores $g_w$ of all feature pairs constitute a fully connected weighted relation graph $G_w$. Note that the edge weight score is degraded to an attention score when $r$ is filled with scalar value 1 (and $\text{diag}(r)$ becomes an entity matrix), and thus it is able to measure weighted feature similarity.

\subsubsection{Static Knowledge Topology.} \label{skt} Although we introduce soft edge weights for all feature pairs, it is also important to globally consider the underlying knowledge of the tabular data. Thus, we use a series of column embeddings to represent the semantics of the tabular features, and a static relation topology score can be computed as follows:
\begin{equation}
    G_t[i,j] = g_t(e_i^h, e_j^t) = \frac{{e_i^h}^T e_j^t}{\Vert e_i^h \Vert_2 \Vert e_j^t \Vert_2}, \label{srs}
\end{equation}
$$e_i^h = E^h \left[:,i\right], \ e_i^t = E^t \left[:,i\right],$$
where $E \in \{E^h, E^t\}$ is learnable column embeddings categorized into the head view or tail view, $E = (e_1, e_2, \dots, e_N) \in \mathds{R}^{d \times N}$, and $d$ is the embedding dimension. Similarly, the relation topology score $g_t$ 
has the symmetric and asymmetric counterparts, and $E^h$ and $E^t$ share parameters in the symmetric relation topology (i.e., $G_t[i,j] \equiv G_t[j,i]$) but are parameter-independent in the asymmetric case (i.e., $G_t[i,j] \ne G_t[j,i]$). We use $L_2$ 
normalization in the $g_t$ score function to transform embeddings to be on a similar scale and improve the training stability.

We generate static relation topology based on the $G_t$ scores in Eq.~(\ref{srs}), as:
\begin{equation}
    A = f_{top}(G_t) = \mathds{1} \left [ \sigma_1(G_t + b) > T \right ], \label{stop}
\end{equation}
where $\sigma_1$ is an element-wise activation parameterised by a learnable bias $b$ (like the operation in PReLU~\cite{he2015delving}), $G_t$ is adjacency matrix scores composed of the relation topology score $g_t$, $T$ is a constant threshold for signal clipping, and $\mathds{1}$ denotes the indicator function. In this way, we obtain a global graph topology (an adjacency matrix $A$) to constrain feature interactions, and this topology can be regarded as static knowledge on the whole task.

\subsection{Relation Graph Assembling} 
As we obtain ``soft'' adaptive edge weights from the data view and ``hard'' static relation graph topology from the knowledge view, we combine them to generate an FR-Graph, following the idea of ``decision on both specific data and underlying knowledge''. Specifically, we assemble the two components as follows:
\begin{equation}
    G = \sigma_2 (f_{\text{nsi}}(A) \odot G_w), \label{frg}
\end{equation}
where $\sigma_2$ is a competitive activation (e.g., $L_p$ normalization, softmax, entmax, sparsemax~\cite{martins2016softmax}) to restrict the indegree of each ``feature node'', and $\odot$ denotes the Hadamard product. The resulted relation graph $G$ is a weighted graph based on both adaptive feature matching and static knowledge topology. To help the FR-Graph focus on learning meaningful interactions between different features, a ``no-self-interaction'' function $f_{\text{nsi}}$ is performed to explicitly exclude self-loops in $G$. We use the FR-Graph to instruct subsequent feature interactions. Since both the edge weights and knowledge topology have the symmetric and asymmetric versions, there are four combinations of FR-Graph covering the complete relation graph. In experiments, we will further discuss the impact of the FR-Graph type.

\section{\ourapproach}

We incorporate GE into the attention-like basic block, and build \ourapproach by stacking multiple blocks for selective tabular feature interaction (see Fig.~\ref{fig-t2g}). \ourapproach uses estimated FR-Graphs to interact features and attain higher-level features layer by layer. A \textit{Cross-level Readout} is sequentially transformed to the feature space of each layer, and selectively collects salient features for the final prediction. A shortcut path is added to preserve the information from the preceding layers, resulted in gated fusion in different feature levels that promotes the model capability.

\subsection{Basic Block} \label{bb}

A single block is built equipped with GE for selective feature interaction (see Fig.~\ref{fig-t2g}(b)). Given input features $X^l \in \mathds{R}^{n \times N}$ to the $l$-th layer, we obtain higher-level features $X^{l+1}$ as follows:
\begin{gather}
    G^l = \textit{GE}(X^l), \ \ V^l = W_v X^l, \label{block} \\
    H^l = G^l V^l + g(X^l), \ \ X^{l+1} = \text{FFN}(H^l) + g(H^l),
\end{gather}
where $W_v \in \mathds{R}^{m \times n}$ is learnable parameters for feature transformation, and $V^l$ is transformed input features. FFN denotes a feed-forward network. As self-interaction is excluded in $G^l$ (see Eq.~(\ref{frg})), a shortcut path $g$ is added to protect the information from the preceding layers, which is a simple \textit{dropout} layer in experiments. Notably, we yield and use the FR-Graph for feature interactions, and does not influence the intra-feature update conducted by the shortcut. In the first layer, we set $X^0$ as the input tabular data encoded by a simple feature tokenizer~\cite{gorishniy2021revisiting}. In this way, higher-level features can be iteratively obtained with the generated FR-Graphs and selective interaction. In implementation, layer normalization is performed (see Fig.~\ref{fig-t2g}(b)) for stable training.

\subsection{Cross-level Readout}

We design a global readout node to selectively collect salient features from each layer and attain comprehensive semantics for the final prediction. Specifically, we attentively fuse selected features at the current layer and combine them with the lower-level features from the preceding layers by a shortcut path. Given the current readout status $z^l \in \mathds{R}^n$, the collection process at the $l$-th layer is defined by:
\begin{gather}
    \alpha_i^l = g_w(h^l, f_i^t) \cdot f_{top}(g_t(e^l, e_i^t)), \ \ h^l = W^h z^l, \\
    r^l = {\rm softmax} (\boldsymbol{\alpha}^l)^T V^l + z^l, \label{readout} \\
    z^{l+1} = \text{FFN}(r^l) + r^l, \label{update}
\end{gather}
where $\alpha_i^l$ denotes the weight of the $i$-th feature that constitutes a weight vector $\boldsymbol{\alpha}^l \in \mathds{R}^N$, $e^l \in \mathds{R}^d$ is a learnable vector representing the semantics of the readout node at the $l$-th layer, $f_i^t$ is an encoded feature (Eq.~(\ref{ft-dg})) of each layer, and $e_i^t$ is a layer-wise column embedding (Eq.~(\ref{srs})). $V^l$ is the transformed input features (Eq.~(\ref{block})). Here we put $z^l$ forward through the same FFN transformation to transform the current readout into the feature space at the ($l+1$)-th layer for the next round of collection. The shortcut paths are directly added without information drop. This collection process is repeated from the input features to the highest-level features, thus encouraging interactions among cross-level features.

\subsection{The Overall Architecture and Training}

Basic blocks are stacked in \ourapproach (Fig.~\ref{fig-t2g}(a)). If without special specification, in experiments we use 8-head \textit{GE} in each block by default (Fig.~\ref{fig-t2g}(b)). Prediction is made based on the readout status after processing the final layer $L$, as:
$$\hat{y} = {\rm FC}({\rm ReLU}({\rm LN}(z^L))),$$
where LN and FC denote layer normalization and a fully connected layer, respectively. As for optimization, we use the cross entropy loss for classification and the mean squared error loss for regression, as in previous DNNs. We tested various tasks and observed that continuing to optimize the static graph topology $A$ in Eq.~(\ref{stop}) across the whole training phase may lead to unstable performance on some easy tasks (e.g., binary classification, small datasets, or few input features). Thus, we freeze it after convergence for further training in a fixed topology manner.

Note that we introduce additional hyperparameters $d$ (Eq.~(\ref{srs})) and $T$ (Eq.~(\ref{stop})). In experiments, we adaptively set $d=2\left \lceil \log_2 N  \right \rceil $ which is for the minimal amount of information to present an adjacency matrix with $N^2$ binary elements, and keep $T = 0.5$ across all the datasets. We choose \textit{sigmoid} as $\sigma_1$ and \textit{softmax} as $\sigma_2$. Straight-through trick~\cite{bengio2013estimating} is used to solve the undifferentiable issue of the indicator function in Eq.~(\ref{stop}).

\section{Experiments}
In this section, we present extensive experimental results and compare with a wide range of state-of-the-art tabular learning DNNs and GBDT. Also, we conduct empirical experiments to examine the impacts of some key \ourapproach components, including comparison of the feature relation graph (FR-Graph) types, ablation study of self-interaction, and the effect of GE. Besides, we explore the model interpretability by visualizing the FR-Graphs and readout selection on two semantically rich datasets.

\subsection{Experimental Setup}

\subsubsection{Datasets.} We use twelve open-source tabular datasets. Gesture Phase Prediction (GE,~\cite{madeo2013gesture}), Churn Modeling (CH, Kaggle dataset), Eye Movements (EY,~\cite{salojarvi2005inferring}), California Housing (CA,~\cite{pace1997sparse}), House 16H (HO, OpenML dataset), Adult (AD,~\cite{kohavi1996scaling}), Helena (HE,~\cite{guyon2019analysis}), Jannis (JA,~\cite{guyon2019analysis}), Otto Group Product Classification (OT, Kaggle dataset), Higgs Small (HI,~\cite{baldi2014searching}), Facebook Comments (FB,~\cite{singh2015comment}), and Year (YE,~\cite{DBLP:conf/ismir/Bertin-MahieuxEWL11}). For each dataset, data preprocessing and train-validation-test splits are fixed for all the methods according to~\cite{gorishniy2021revisiting, gorishniy2022embeddings}. Dataset statistics are given in Table~\ref{dataset-detail}, and more details are in Appendix~\ref{data-stat}.

\begin{table*}[t]
\centering
\begin{tabular}{lcccccccccccc}
\toprule
Dataset& GE& CH& EY& CA& HO& AD& OT& HE& JA& HI& FB& YE\\
\midrule
\# features& 32& 9+1& 26& 8& 16& 6+8& 93& 27& 54& 28& 50+1& 90\\
\# samples& 9873& 10000& 10936& 20640& 22784& 48842& 61878& 65196& 83733& 98050& 197080& 515345\\
\# classes& 5& 2& 3& -& -& 2& 9& 100& 4& 2& -& -\\
Metric& Acc. &Acc.& Acc.& RMSE& RMSE& Acc.& Acc.& Acc.& Acc.& Acc.& RMSE& RMSE\\
\bottomrule
\end{tabular}

\caption{Some details of the 12 public datasets. "RMSE" denotes root mean squared error (for regression), and ``Acc.'' means accuracy (for classification). The number following each ``+'' in the row of ``\# features'' is the number of categorical features.}\label{dataset-detail}
\end{table*}

\begin{table*}[t]
\centering
\begin{tabular}{l|cccccccccccc}
\toprule
 & GE $\uparrow$& CH $\uparrow$& EY $\uparrow$& CA $\downarrow$& HO $\downarrow$& AD $\uparrow$& OT $\uparrow$& HE $\uparrow$& JA $\uparrow$& HI $\uparrow$& FB $\downarrow$& YE $\downarrow$\\
\midrule
XGBoost & 68.42	&85.92	&72.51	&0.436	&3.169	&87.30	&82.46	&37.47	&71.85	&72.41	&5.359	&8.850\\
\midrule
MLP& 58.64	&85.77	&61.10	&0.499	&3.173	&85.35	&80.99	&38.38	&71.97	&72.00	&5.943	&8.849\\
SNN& \underline{64.69}	&85.74	&61.55	&0.498	&3.207	&85.40	&81.17	&37.19	&71.94	&72.21	&5.892	&8.901\\
TabNet& 60.01	&85.01	&62.08	&0.513	&3.252	&84.84	&79.06	&37.86	&72.26	&71.97	&6.559	&8.916\\
DANet-28& 61.63	&85.10	&60.53	&0.524	&3.236	&85.00	&81.04	&35.45	&70.72	&71.47	&6.167	&8.914\\
NODE& 53.94	&85.86	&65.54	&0.463	&3.216	&\underline{85.77}	&80.37	&35.33	&72.78	&72.51	&\textbf{5.698}	&\textbf{8.777}\\
AutoInt& 58.33	&85.51	&61.07	&0.472	&3.147	&85.66	&80.11	&37.26	&72.08	&72.51	&5.852	&8.862\\
DCNv2& 55.72	&85.68	&61.37	&0.489	&3.172	&85.48	&80.15	&38.61	&71.56	&72.20	&5.847	&8.882\\
FT-Transformer& 61.25	&\underline{86.07}	&\underline{70.84}	&\underline{0.460}	&\textbf{3.124}	&85.72	&\underline{81.30}	&\textbf{39.10}	&\underline{73.24}	&\underline{73.06}	&6.079	&8.852\\
\ourapproach& \textbf{65.57}	&\textbf{86.25}	&\textbf{78.18}	&\textbf{0.455}	&\underline{3.138}	&\textbf{85.96}	&\textbf{81.87}	&\underline{39.06}	&\textbf{73.68}	&\textbf{73.39}	&\underline{5.701}	&\underline{8.851}\\
\bottomrule
\end{tabular}

\caption{Performance comparison on the 12 public tubular datasets. Each result reported is averaged over 15 random seeds. For standard deviations, see Appendix~\ref{exp-detail}. For each dataset, the top performances among the DNNs are marked in \textbf{bold}, and the second best results are \underline{underlined}. We also report XGBoost results as a typical representation of GBDT models. $\downarrow$ represents the RMSE metric (the lower the better) and $\uparrow$ represents accuracy (the higher the better).}\label{main-result}
\end{table*}

\subsubsection{Implementation Details.} We implement our \ourapproach model using PyTorch on Python 3.8. All the experiments are run on NVIDIA RTX 3090. In training, if without special specification, we use FR-Graphs with symmetric edge weights and asymmetric graph topology in GE. The optimizer is AdamW~\cite{loshchilov2018decoupled} with the default configuration except for the learning rate and weight decay rate. For DANet-28, we follow its QHAdam optimizer~\cite{ma2018quasi} and the pre-set hyperparameters given in~\cite{chen2022danets} without tuning. For the other DNNs and XGBoost, we follow the settings provided in~\cite{gorishniy2021revisiting} (including the optimizers and hyperparameter spaces), and perform hyperparameter tuning with the Optuna library~\cite{akiba2019optuna} and grid search (only for NODE). More detailed information of hyperparameters is provided in Appendix~\ref{hyper-tune}.

\subsubsection{Comparison Methods.} 
In our 
experiments, we compare our \ourapproach with the representative non-deep method XGBoost~\cite{chen2016xgboost} and the known DNNs, including NODE~\cite{popov2019neural}, AutoInt~\cite{song2019autoint}, TabNet~\cite{arik2021tabnet}, DCNv2~\cite{wang2021dcn}, FT-Transformer~\cite{gorishniy2021revisiting}, and DANets~\cite{chen2022danets}. Some other common DNNs such as MLP and SNN (an MLP network with SELU activation)~\cite{klambauer2017self} are taken into comparison as well.

\subsection{Main Results and Analyses}

\subsubsection{Performance Comparison.} The performances of the DNNs and non-deep models are reported in Table~\ref{main-result}. \ourapproach outperforms these DNNs on eight datasets, and is comparable with XGBoost in most the cases. All the models are hyperparameter-tuned by choosing the best validation results with Optuna-driven tuning~\cite{akiba2019optuna}.

\subsubsection{The Effect of FR-Graph Types.} We compare four types of FR-Graphs in GE.  Table~\ref{frg-comparison} reports the results, from which one can see that it is often better to choose symmetric edge weights and asymmetric knowledge topology. This suggests that mutual interactions between two tabular features are likely to be the same, and asymmetric topology offers a larger semantic exploration space that is more likely to yield useful features. The results on the other datasets are provided in Appendix~\ref{exp-detail}.

\begin{table}[t]
\centering
\resizebox{.95\columnwidth}{!}{
\begin{tabular}{c|cccccc}
\toprule
FR-Graph & EY $\uparrow$& CA ($\times$100) $\downarrow$& HO $\downarrow$& OT $\uparrow$& FB $\downarrow$& YE $\downarrow$\\
\midrule
${\rm A^wS^t}$ &77.34	&45.73	&3.171   &81.80   &5.736	&8.886\\
\midrule
${\rm A^wA^t}$ &77.59	&45.77	&3.145   &81.85   &5.718	&8.861\\
\midrule
${\rm S^wS^t}$ &76.46	&45.61	&3.151    &81.89   &5.723	&8.885\\
\midrule
${\rm S^wA^t}$ (ours) &78.18	&45.53	&3.138   &81.87   &5.701	&8.851\\
\bottomrule
\end{tabular}
}

\caption{Comparison of four FR-Graph types on several tasks and datasets. ``A'' means asymmetric, and ``S'' means symmetric. ``${\rm A^wS^t}$'', for example, is for asymmetric edge weights and symmetric graph topology. Likewise, ``${\rm A^wA^t}$'', ``${\rm S^wS^t}$'', and ``${\rm S^wA^t}$'' denote the other three types of FR-Graphs. The performances on the CA dataset are scaled ($\times$100) in these several tables for more clear comparison.
}\label{frg-comparison}
\end{table}

\subsubsection{The Effect of Self-interaction.} One of our key designs in GE is the ``no self-interaction function'' that explicitly excludes self-loops in FR-Graphs. Table~\ref{nsl-comparison} reports comparison results on several datasets with no self-loop FR-Graphs (ours) and self-loop FR-Graphs. The results show that in most the cases, removing self-loops and focusing on interactions with other features slightly benefit performances in both classification and regression. This may be because feature self-interaction affects the probabilities of interactions with other features (as we use competitive activation in Eq.~(\ref{frg})), while our shortcut paths have already preserved self-information.

\begin{table}[t]
\centering
\resizebox{.95\columnwidth}{!}{
\begin{tabular}{c|cccccc}
\toprule
 & EY $\uparrow$& CA ($\times$100) $\downarrow$& HO $\downarrow$& OT $\uparrow$& FB $\downarrow$& YE $\downarrow$\\
\midrule
w/o SL (ours) &78.18	&45.53	&3.138   &81.87   &5.701	&8.851\\
\midrule
SL &77.89	&45.62	&3.152	&81.81	&5.691	&8.856\\
\midrule
SL $-$ w/o SL &$-0.29$	&0.09	&0.014	&$-0.06$	&$-0.01$	&0.005\\
\bottomrule
\end{tabular}
}

\caption{Comparison of the effects of FR-Graphs without (w/o) self-loops and FR-Graphs with self-loops. ``SL'' means self-loops.}\label{nsl-comparison}
\end{table}

\subsubsection{The Effect of GEs.} We explore the impact of including GEs at different layers of \ourapproach. Table~\ref{ge-pos-comparison} reports the performances of different model versions which differ solely in the positions and numbers of GEs used. As for the layers without GE, we use the ordinary attention score for substitution. Overall, the positions of GEs show bigger influence on regression tasks than on classification tasks. As one can see, in regression tasks, the model incurs larger performance drops when GEs are equipped in higher layers, while the drops do not seem so large related to GE positions in classification. Also, the model equipped with only attention score is better than the one with a single GE in a high layer (not in the first layer) for regression tasks, but is always sub-optimal in classification tasks. 
A probable explanation is that regression needs a smoother optimization space than classification, and thus the fully connected attention score provides the kind of interactions to cope with continuous feature values, while a single GE in a high layer is difficult to capture clear relations among features fused in the fully connected manner. Therefore, it is better to completely use attention score than a single GE in a high layer for regression. A single GE in the first layer shows the least performance drop in both regression and classification, which can be explained by the strength of GE in capturing underlying relations among tabular features with clear semantics. 

In summary, the removal of GE in any layers is likely to cause performance drop, and the best results are achieved by applying GE to all the layers.

\begin{table}[t]
\centering
\resizebox{.55\columnwidth}{!}{
\begin{tabular}{c|cccccc}
\toprule
 & CA ($\times$100) $\downarrow$& JA $\uparrow$\\
\midrule
All &45.53	&73.68\\
\midrule
\# 1 &45.78 ($+0.25$)	&73.40 ($-0.28$)\\
\midrule
\# 2 &45.96 ($+0.43$)	&73.31 ($-0.37$)\\
\midrule
\# 3 &46.06 ($+0.53$)	&73.37 ($-0.31$)\\
\midrule
None &45.84 ($+0.31$)	&73.23 ($-0.45$)\\
\bottomrule
\end{tabular}
}

\caption{Performances of including GEs in different layers of \ourapproach. All the results are obtained with a 3-layer \ourapproach. ``\# $i$'' means that only the $i$-th layer has GE while the other layers replace GE with the ordinary attention score, ``All'' means that all the layers are equipped with GE, and ``None'' means that all the layers use ordinary attention.}\label{ge-pos-comparison}
\end{table}

\subsubsection{Comparison of Topology Learning Approaches.} Apart from the column embedding approach proposed in Sec.~\ref{skt}, there are some other intuitive straightforward approaches to get knowledge topology of the RF-Graph, for example, performing threshold clipping on the adaptive edge weights directly (we call it ``\textbf{adaptive} topology'') or learning an $N$-by-$N$ adjacency matrix (we call it ``\textbf{free} topology''). Concretely, for learning \textbf{adaptive} topology, we substitute $G_t$ in Eq.~(\ref{stop}) with $G_w$ in Eq.~(\ref{dg}). For learning the \textbf{free} topology, we directly represent $G_t$ by an $N$-by-$N$ matrix. Table~\ref{top-strategy} reports the comparison results of these topology learning strategies. One can see that, the static knowledge topology shared on the whole dataset (our approach) attains superior performances than the adaptive topology, implying the plausibility of our underlying knowledge assumption mentioned in Sec.~\ref{intro}. Besides, the completely free topology also achieves inferior performances, which is probably because of the excessive freedom given to the learnable matrix.

\begin{table}[t]
\centering
\resizebox{.95\columnwidth}{!}{
\begin{tabular}{c|cccccc}
\toprule
Topology & CA ($\times$100) $\downarrow$& JA $\uparrow$ & Complexity & Fixed \\
\midrule
ours &45.53	&73.68 & O($N$log$N$) & Y\\
\midrule
adaptive &45.88 ($+0.35$)	&73.08 ($-0.60$) & O(1) & N\\
\midrule
free &45.87 ($+0.34$)	&73.46 ($-0.22$) & O($N^2$) & Y\\
\bottomrule
\end{tabular}
}

\caption{Performances of different topology learning approaches. ``Complexity'' indicates the additional space computational complexity (the amount of extra model parameters) caused by the number of tabular features $N$. ``Fixed'' indicates whether the learned topology is data-adaptive (N) or static (Y).}\label{top-strategy}
\end{table}

\subsubsection{Comparison with DANet Grouped Interactions.} As illustrated in Fig.~\ref{fig-illustration}, DANets~\cite{chen2022danets} interacted tabular features in the group determined by the ``entmax'' operation. Here we compare our \textbf{graph-based} interaction with that \textbf{group-based} one to inspect the benefits of FR-Graph. Specifically, we substitute the knowledge topology $A$ in Eq.~(\ref{stop}) with DANet grouped selection mask.
The results in Table~\ref{group-interaction} suggest that it of greater benefits to organize tabular features into a graph, since a graph topology is able to capture relation edges and provide more subtle interactions than a group structure.

\begin{table}[t]
\centering
\resizebox{.95\columnwidth}{!}{
\begin{tabular}{c|cccccc}
\toprule
Interaction & CA ($\times$100) $\downarrow$ & HO $\downarrow$ & JA $\uparrow$ \\
\midrule
graph (ours) &45.53 & 3.138	&73.68\\
\midrule
group (DANet) &45.88 ($+0.35$) & 3.215 ($+0.077$)	&73.08 ($-0.60$)\\
\bottomrule
\end{tabular}
}

\caption{Comparison with DANet group-based interaction on several datasets.}\label{group-interaction}
\end{table}

\begin{figure}[t]
\centering
\includegraphics[width=0.5\textwidth]{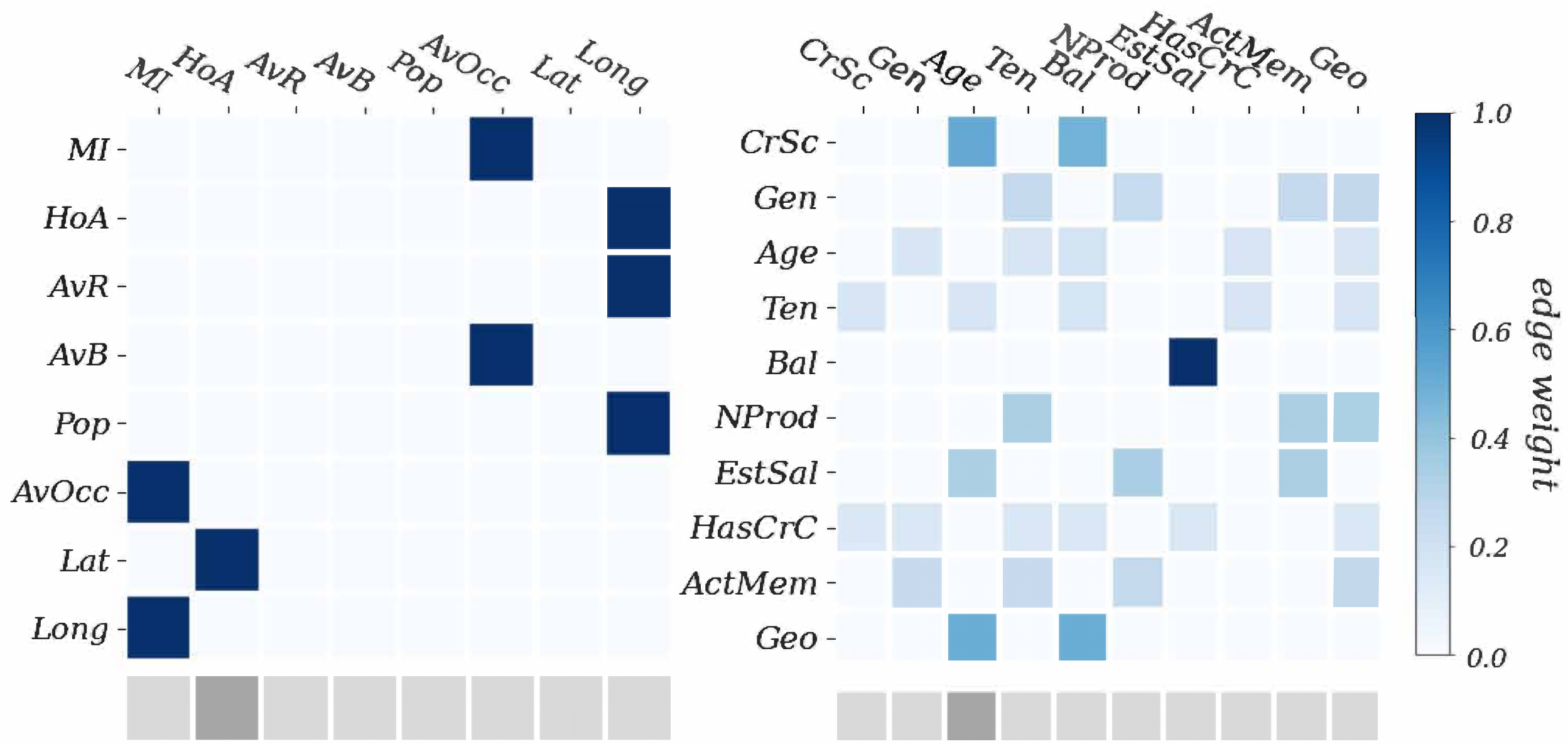} 

\caption{Visualization of the FR-Graph in the first layer (heat map) and the readout selection (dark bar) on the datasets CA (left) and CH (right). More details of the feature descriptions are given in Appendix~\ref{feat-description}.}
\label{fig-interpret}

\end{figure}

\subsection{Interpretability}

In Fig.~\ref{fig-interpret}, we visualize the first-layer FR-Graph and the readout collecting strategy on the input features (i.e., features from the feature tokenizer; see Fig.~\ref{fig-t2g}(a)). On the CA dataset, it is reasonable to find that the median income (\textit{MedInc}, \textit{MI}) of the residents within a block group is related to the average number of the household members (\textit{AveOccup}, \textit{AvOcc}), and \textit{AveOccup} can affect the average number of bedrooms (\textit{AveBedrooms}, \textit{AvB}). Also, there appear to be some relations such as \textit{Longitude} (\textit{Long})-\textit{HouseAge} (\textit{HoA}), \textit{Longitude}-\textit{AveRooms} (\textit{AvR}), and \textit{Longitude}-\textit{Population} (\textit{Pop}), which are probably derived from dataset bias. As for readout, one can see that solely \textit{HouseAge} is collected that is a meaningful feature in house price prediction. On the CH dataset, there are reasonable relations between \textit{Balance} (\textit{Bal}, bank balance of a customer) and \textit{EstimatedSalary} (\textit{EstSal}), as well as the age of the customer (\textit{Age}) and \textit{EstimatedSalary}. Also, it is interpretable that the credit score of a customer (\textit{CreditScore}, \textit{CrSc}) is highly related to that customer's \textit{Age} and \textit{Balance}. The readout collects only \textit{Age} in the current level for predicting whether a customer will leave the bank, which is intuitive as well.
\section{Conclusions}

In this paper, we proposed \ourapproach, a new bespoke Transformer model for tabular learning with a novel module \textit{Graph Estimator} (GE) for promoting heterogeneous feature interaction based on estimated relation graphs. We adapted feature relation graphs into the basic blocks of \ourapproach in an attention-like fashion for simplicity and applicability. Experiments on extensive public datasets showed that \ourapproach achieves better performances than various DNNs and is comparable with XGboost. We expect that our \ourapproach will serve as a strong baseline in tabular learning studies and enhance research interest in handling feature heterogeneity of tabular data.
\section*{Acknowledgments}

This research was partially supported by the National Key R\&D Program of China under grant No. 2018AAA0102102 and National Natural Science Foundation of China under grants No. 62132017.

\bibliography{aaai23}

\newpage
\appendix

\section{Dataset Descriptions} \label{data-stat}

Details of used datasets are shown in Table~\ref{datasets}. We follow the same train-valid-test splits and data pre-processing methods in~\cite{gorishniy2021revisiting, gorishniy2022embeddings}.

\begin{table*}[t]
\centering
\begin{tabular}{lcccccccccccc}
\toprule
Dataset& \# Train& \# Validation& \# Test& \# Num& \# Cat& Task& Batch size\\
\midrule
Gesture Phase(GE)& 6318& 1580& 1795& 32& 0& Multiclass& 128\\
Churn Modelling(CH)& 6400& 1600& 2000& 9& 1& Binclass& 128\\
Eye Movements(eye)& 6998& 1750& 2188& 26& 0& Multiclass& 128\\
California Housing(CA)& 31209&3303& 4128& 8& 0& Regression& 256\\
House 16H(HO)& 14581&3646& 4557& 16& 0& Regression& 256\\
Adult(AD)& 26048&6513& 16281& 6& 8& Binclass& 256\\
Otto Group Products(OT)& 39601&9901& 12376& 93& 0& Multiclass& 512\\
Helena(HE)& 41724&10432& 13040& 27& 0& Multiclass& 512\\
Jannis(JA)& 53588&13398& 16747& 54& 0& Multiclass& 512\\
Higgs Small(HI)& 62752&15688& 19610& 28& 0& Binclass& 512\\
Facebook Comments Volume(FB)& 157638&19722& 19720& 50& 1& Regression& 512\\
Year(YE)& 370972&92743& 51630& 90& 0& Regression& 1024\\
\bottomrule
\end{tabular}

\caption{Dataset details. ``\# Num'' and ``\# Cat'' mean the numbers of numerical features and categorical features in each dataset.}\label{datasets}
\end{table*}

\section{Hyperparameter Tuning} \label{hyper-tune}

For baseline architectures of XGBoost, MLP, SNN, TabNet, NODE, AutoInt, DCNv2 and FT-Transformer, we reuse the implementations and hyperparameter search spaces in~\cite{gorishniy2021revisiting} for comparison. For DANets we use 28-layer architecture and corresponding pre-set hyperparameters as~\cite{chen2022danets} recommended without tuning. We use grid search for NODE. For our \ourapproach, we use the optuna-driven tuning~\cite{akiba2019optuna} with hyperparameter spaces reported in Table \ref{hyper}. we also use the optuna-driven tuning~\cite{akiba2019optuna} for the rest methods.

\begin{table*}[ht]
\centering
\begin{tabular}{ll}
\toprule
Parameter& Distribution\\
\midrule
\# layers & (A) UniformInt$\left [ 1,5 \right ]$, (B) UniformInt$\left [ 1,3 \right ]$\\
Feature embedding size & (A,B) UniformInt$\left [ 64,512 \right ]$\\
Residual Dropout & (A) Uniform$\left [ 0,0.2 \right ]$, (B) Const(0.0)\\
Attention Dropout & (A,B) Uniform$\left [ 0,0.5 \right ]$\\
FNN Dropout & (A,B) Uniform$\left [ 0,0.5 \right ]$\\
Learning rate (main backbone) & (A) LogUniform$\left [ 1e-5,1e-3 \right ]$, (B) LogUniform$\left [ 3e-5,3e-4 \right ]$\\
Learning rate (column embedding) & (A,B) LogUniform$\left [ 5e-3,5e-2 \right ]$\\
Weight decay & (A,B) LogUniform$\left [ 1e-6,1e-3 \right ]$\\
\midrule
\# iterations & 100\\
\bottomrule
\end{tabular}

\caption{Hyperparameter tuning spaces for \ourapproach. (A) = \{FB, YE\}, (B) = \{GE, CH, EY, CA, HO, AD, OT, HE, JA, HI\}.}\label{hyper}
\end{table*}

\section{Detailed Experiment Results} \label{exp-detail}
Table \ref{main-result-detail} reports detailed performance results (with standard deviations) of all models on all the datasets. Table \ref{frg-comparison-detail} reports detailed results in ``FR-Graph combinations''.

\begin{sidewaystable}[t]
\centering
\resizebox{.95\columnwidth}{!}{
\begin{tabular}{lcccccccccccc}
\toprule
 & GE $\uparrow$& CH $\uparrow$& EY $\uparrow$& CA $\downarrow$& HO $\downarrow$& AD $\uparrow$& OT $\uparrow$& HE $\uparrow$& JA $\uparrow$& HI $\uparrow$& FB $\downarrow$& YE $\downarrow$\\
\midrule
XGBoost & 68.42$\pm$0.51	&85.92$\pm$0.10	&72.51$\pm$0.51	&0.436$\pm$1.40e-6	&3.169 $\pm$1.45e-4	&87.30$\pm$4.31e-3	&82.46$\pm$0.07	&37.47$\pm$0.13	&71.85$\pm$0.06	&72.41$\pm$0.02	&5.359$\pm$8.20e-4	&8.850$\pm$3.36e-5\\
\midrule
MLP& 58.64$\pm$3.18	&85.77$\pm$0.05	&61.10$\pm$0.67	&0.499$\pm$1.90e-5	&3.173$\pm$8.03e-4	&85.35$\pm$0.02	&80.99$\pm$0.22	&38.38$\pm$0.04	&71.97$\pm$0.05	&72.00$\pm$0.05	&5.943$\pm$3.90e-2	&8.849$\pm$1.20e-2\\
SNN& \underline{64.69$\pm$0.99}	&85.74$\pm$0.10	&61.55$\pm$1.36	&0.498$\pm$7.09e-5	&3.207$\pm$6.89e-4	&85.40$\pm$0.02	&81.17$\pm$0.06	&37.19$\pm$0.12	&71.94$\pm$0.10	&72.21$\pm$0.05	&5.892$\pm$3.56e-2	&8.901$\pm$5.74e-4\\
TabNet& 60.01$\pm$0.61	&85.01$\pm$1.30	&62.08$\pm$0.93	&0.513$\pm$1.61e-4	&3.252$\pm$3.38e-3 &84.84$\pm$0.39	&79.06$\pm$0.17	&37.86$\pm$0.08	&72.26$\pm$0.08	&71.97$\pm$0.05	&6.559$\pm$4.56e-2	&8.916$\pm$7.98e-4\\
DANet-28& 61.63$\pm$1.46	&85.10$\pm$0.20	&60.53$\pm$1.51	&0.524$\pm$5.25e-5	&3.236$\pm$3.96e-3	&85.00$\pm$0.08	&81.04$\pm$0.02	&35.45$\pm$0.06	&70.72$\pm$0.05	&71.47$\pm$0.05	&6.167$\pm$9.31e-2	&8.914$\pm$2.97e-3\\
NODE& 53.94$\pm$1.17	&85.86$\pm$0.13	&65.54$\pm$0.44	&0.463$\pm$2.60e-6	&3.216$\pm$4.65e-5	&\underline{85.77$\pm$0.03}	&80.37$\pm$0.04	&35.33$\pm$0.22	&72.78$\pm$0.01	&72.51$\pm$0.01	&\textbf{5.698$\pm$1.59e-2}	&\textbf{8.777$\pm$1.43e-4}\\
AutoInt& 58.33$\pm$2.13	&85.51$\pm$0.10	&61.07$\pm$0.69	&0.472$\pm$1.76e-5	&3.147$\pm$2.90e-4	&85.66$\pm$0.02	&80.11$\pm$0.06	&37.26$\pm$0.03	&72.08$\pm$0.02	&72.51$\pm$0.04	&5.852$\pm$3.67e-2	&8.862$\pm$7.23e-4\\
DCNv2& 55.72$\pm$1.73	&85.68$\pm$0.02	&61.37$\pm$0.32	&0.489$\pm$1.51e-5	&3.172$\pm$7.96e-4	&85.48$\pm$0.14	&80.15$\pm$0.21	&38.61$\pm$0.09	&71.56$\pm$0.02	&72.20$\pm$0.03	&5.847$\pm$2.41e-2	&8.882$\pm$8.13e-4\\
FT-Transformer& 61.25$\pm$5.82	&\underline{86.07$\pm$0.08}	&\underline{70.84$\pm$1.18}	&\underline{0.460$\pm$1.13e-5}	&\textbf{3.124$\pm$1.35e-3}	&85.72$\pm$0.04	&\underline{81.30$\pm$0.09}	&\textbf{39.10$\pm$0.05}	&\underline{73.24$\pm$0.04}	&\underline{73.06$\pm$0.03}	&6.079$\pm$5.00e-2	&8.852$\pm$5.32e-4\\
\ourapproach& \textbf{65.57$\pm$1.44}	&\textbf{86.25$\pm$0.05}	&\textbf{78.18$\pm$1.85}	&\textbf{0.455$\pm$2.10e-5}	&\underline{3.138$\pm$2.51e-3}	&\textbf{85.96$\pm$0.02}	&\textbf{81.87$\pm$0.03}	&\underline{39.06$\pm$0.03}	&\textbf{73.68$\pm$0.03}	&\textbf{73.39$\pm$0.03}	&\underline{5.701$\pm$1.61e-2}	&\underline{8.851$\pm$1.32e-3}\\
\bottomrule
\end{tabular}
}

\caption{Detailed results with standard deviations of all models.}\label{main-result-detail}
\end{sidewaystable}

\begin{sidewaystable}[ht]
\centering
\resizebox{.95\columnwidth}{!}{
\begin{tabular}{c|cccccccccccc}
\toprule
Assemblings & GE $\uparrow$& CH $\uparrow$& EY $\uparrow$& CA $\downarrow$& HO $\downarrow$& AD $\uparrow$& OT $\uparrow$& HE $\uparrow$& JA $\uparrow$& HI $\uparrow$& FB $\downarrow$& YE $\downarrow$\\
\midrule
${\rm A^wS^t}$ & 65.43$\pm$1.34&    85.25$\pm$0.03 &77.34$\pm$2.54	&0.457$\pm$1.93e-5	&3.171$\pm$2.42e-3 &85.95$\pm$0.01	  &81.80$\pm$0.02  &39.09$\pm$0.04 &73.68$\pm$0.03 &73.49$\pm$0.02 &5.736$\pm$4.67e-3 &8.886$\pm$9.59e-4  \\
\midrule
${\rm A^wA^t}$ &65.11$\pm$1.51  &86.26$\pm$0.05 &77.59$\pm$2.56	&0.458$\pm$2.27e-5	&3.145$\pm$2.67e-3    &85.98$\pm$0.01	&81.85$\pm$0.02  &39.02$\pm$0.03   &73.70$\pm$0.03 &73.35$\pm$0.05 &5.718$\pm$3.40e-3   &8.861$\pm$1.62e-3  \\
\midrule
${\rm S^wS^t}$ &64.90$\pm$2.88  &86.20$\pm$0.02 &76.46$\pm$4.67	&0.456$\pm$1.22e-5	&3.151$\pm$2.45e-3  &85.97$\pm$0.02	  &81.89$\pm$0.03    &39.17$\pm$0.01 &73.78$\pm$0.02 &73.69$\pm$0.05 &5.723$\pm$3.36e-3   &8.885$\pm$2.83e-3 \\
\midrule
${\rm S^wA^t}$(ours) & 65.57$\pm$1.44	& 86.25$\pm$0.05	& 78.18$\pm$1.85 & 0.455$\pm$2.10e-5 	& 3.138$\pm$2.51e-3	& 85.96$\pm$0.02 & 81.87$\pm$0.03	& 39.06$\pm$0.03	& 73.68$\pm$0.03	& 73.39$\pm$0.03	& 5.701$\pm$1.61e-2	& 8.851$\pm$1.32e-3 \\
\bottomrule
\end{tabular}
}

\caption{Comparison details of four FR-Graph assemblings.}\label{frg-comparison-detail}
\end{sidewaystable}

\section{Dataset Features} \label{feat-description}
Table \ref{feature-semantic} shows detailed descriptions of features in dataset CA and CH. The CA dataset is typically used for predicting house price in California, while the CH is often used for predicting whether the customer will leave the bank.

\begin{table*}[ht]
\centering
\begin{tabular}{c|lll}
\toprule
Dataset& Feature& Abbr& Description\\
\midrule
\multirow{8}{*}{CA} & \textit{MedInc} & \textit{MI} & Median income within a block.\\
 & \textit{HouseAge} & \textit{HoA} & Median house age within a block.\\
 & \textit{AveRooms} & \textit{AvR} & Average number of rooms per household.\\
 & \textit{AveBedrms} & \textit{AvB} & Average number of bedrooms per household.\\
 & \textit{Population} & \textit{Pop} & Total number of people residing within a block.\\
 & \textit{AveOccup} & \textit{AvOcc} & Average number of household members.\\
 & \textit{Latitude} & \textit{Lat} & Block group latitude.\\
 & \textit{Longitude} & \textit{Long} & Block group longitude.\\
\midrule
\multirow{10}{*}{CH}& \textit{CreditScore} & \textit{CrSc} & Credit score of the customer.\\
 & \textit{Gender} & \textit{Gen} & Gender of the customer.\\
 & \textit{Age} & \textit{Age} & Age of the customer.\\
 & \textit{Tenure} & \textit{Ten} & Number of years for which the customer has been with the bank.\\
 & \textit{Balance} & \textit{Bal} & Bank balance of the customer.\\
 & \textit{NumOfProducts} & \textit{NProd} & Number of bank products the customer is utilising.\\
 & \textit{EstimatedSalary} & \textit{EstSal} & Estimated salary of the costumer.\\
 & \textit{HasCrCard} & \textit{HasCrC} & Whether the customer has a credit card in the bank.\\
 & \textit{IsActiveMember} & \textit{ActMem} & Whether the customer is an active member in the bank.\\
 & \textit{Geography} & \textit{Geo} & The country from which the customer belongs.\\
\bottomrule
\end{tabular}

\caption{Feature descriptions of California Housing (CA) and Churn Modeling (CH).}\label{feature-semantic}
\end{table*}

\end{document}